\documentclass[lettersize,journal]{IEEEtran}
\usepackage{amsmath,amsfonts}
\usepackage{algorithmic}
\usepackage{algorithm}
\usepackage{array}
\usepackage[caption=false,font=normalsize,labelfont=sf,textfont=sf]{subfig}
\usepackage{textcomp}
\usepackage{stfloats}
\usepackage{url}
\usepackage{verbatim}
\usepackage{graphicx}
\usepackage{cite}
\usepackage{caption}
\usepackage{hyperref}
\makeatletter
\renewcommand{\maketag@@@}[1]{\hbox{\m@th\normalsize\normalfont#1}}%
\makeatother

\begin{document}

\title{Dual-stream Transformer-GCN Model with Contextualized Representations Learning for Monocular 3D Human Pose Estimation}

\author{Mingrui Ye, 
        Lianping Yang,
        Hegui Zhu,
        Zenghao Zheng,
        Xin Wang,
        Yantao Lou
        
\thanks{This paper was produced by the Applied Mathematics Group of Northeastern University. This work is supported by the National Natural Science Foundation of China (No. 62472080) and by Liaoning Provincial Natural Science Foundation (2023-MS-330). Corresponding author: Lianping Yang.

Mingrui Ye, Lianping Yang, Hegui Zhu and Zenghao Zheng are with the College of Sciences, Northeastern University, Shengyang 110819, China (email: 2100145@stu.neu.edu.cn, yanglp@mail.neu.edu.cn, zhuhegui@mail.neu.edu.cn, 2300178@stu.neu.edu.cn)

Mingrui Ye is also with the department of informatics, King's College London, London, UK (email: mingrui.ye@kcl.ac.uk)

Lianping Yang, Hegui Zhu are also with Key Laboratory of Differential Equations and Their Applications, Northeastern University, Liaoning Provincial Department of Education

Xin Wang and Yantao Lou are with Shenyang sport University, Shenyang, China

}
}



\maketitle

\begin{abstract}
This paper introduces a novel approach to monocular 3D human pose estimation using contextualized representation learning with the Transformer-GCN dual-stream model. Monocular 3D human pose estimation is challenged by depth ambiguity, limited 3D-labeled training data, imbalanced modeling, and restricted model generalization. To address these limitations, our work introduces a groundbreaking motion pre-training method based on contextualized representation learning. Specifically, our method involves masking 2D pose features and utilizing a Transformer-GCN dual-stream model to learn high-dimensional representations through a self-distillation setup. By focusing on contextualized representation learning and spatial-temporal modeling, our approach enhances the model's ability to understand spatial-temporal relationships between postures, resulting in superior generalization. Furthermore, leveraging the Transformer-GCN dual-stream model, our approach effectively balances global and local interactions in video pose estimation. The model adaptively integrates information from both the Transformer and GCN streams, where the GCN stream effectively learns local relationships between adjacent key points and frames, while the Transformer stream captures comprehensive global spatial and temporal features. Our model achieves state-of-the-art performance on two benchmark datasets, with an MPJPE of 38.0mm and P-MPJPE of 31.9mm on Human3.6M, and an MPJPE of 15.9mm on MPI-INF-3DHP. Furthermore, visual experiments on public datasets and in-the-wild videos demonstrate the robustness and generalization capabilities of our approach. The code is available at {\url{https://github.com/bigrayss/Motion2vec}}.
\end{abstract}


\section{Introduction}
\IEEEPARstart{3}{D} human pose estimation (HPE), which has attracted considerable attention in academia and industry, aims to extract 3D pose geometry and motion information from input data captured by various sensors. It finds extensive applications in motion analysis, human-computer interaction, augmented reality (AR), virtual reality (VR), and healthcare. Among the various methods, monocular 3D HPE stands out due to its reliance on RGB images and videos from a single perspective, obviating the need for multiple viewpoints. This approach reduces deployment costs and is suitable for numerous practical scenarios, such as monitoring systems and smartphone applications. Furthermore, monocular methods offer a feasible foundation for multi-view research.

However, monocular 3D HPE with deep learning faces several challenges, including depth ambiguity, occlusion, distortion, insufficient 3D-labeled training data, imbalanced modeling, and constrained model generalization. Addressing these issues requires the use of video data to provide additional sequential information. Consequently, effectively leveraging spatial-temporal information becomes crucial for successful monocular 3D HPE tasks.

Inspired by the success of the transformer \cite{vaswani2017attention} in sequence modeling, numerous 2D-to-3D two-stage methods have leveraged transformer encoders for pose estimation. Existing works \cite{zheng20213d, zhang2022mixste,tang20233d, sun2024repose} utilize transformer blocks to encode temporal and spatial information separately. However, these methods suffer from high computational complexity and weak local awareness. Other scholars primarily approach HPE tasks from a temporal perspective \cite{li2022exploiting, shan2022p}, conducting temporal interactions for long-range frames and gathering information hierarchically using temporal convolutional networks (TCN) and attention mechanisms. Although these methods can achieve promising accuracy with low computational costs, their spatial awareness is limited since they overlook the differences in information between key points.

Alternatively, studies such as \cite{wang2020motion, hu2021conditional} combine spatial-temporal dependencies using spatial graph convolution networks (GCN) and TCN separately, proposing the use of U-Net to capture both short-term and long-term temporal dependencies. However, these approaches diminish the network's ability to model the 3D pose in sequences compared to transformer-based methods. Moreover, \cite{shan2022p, lin2021end, zhu2023motionbert} recognize that learning temporal-spatial correlations is challenging, as optimizing the model directly from 2D to 3D is difficult. Consequently, they propose various pre-training methods. Nevertheless, these pre-training methods also encounter issues in feature alignment.

\begin{figure*}[!t]
  \centering
  \includegraphics[scale=0.4]{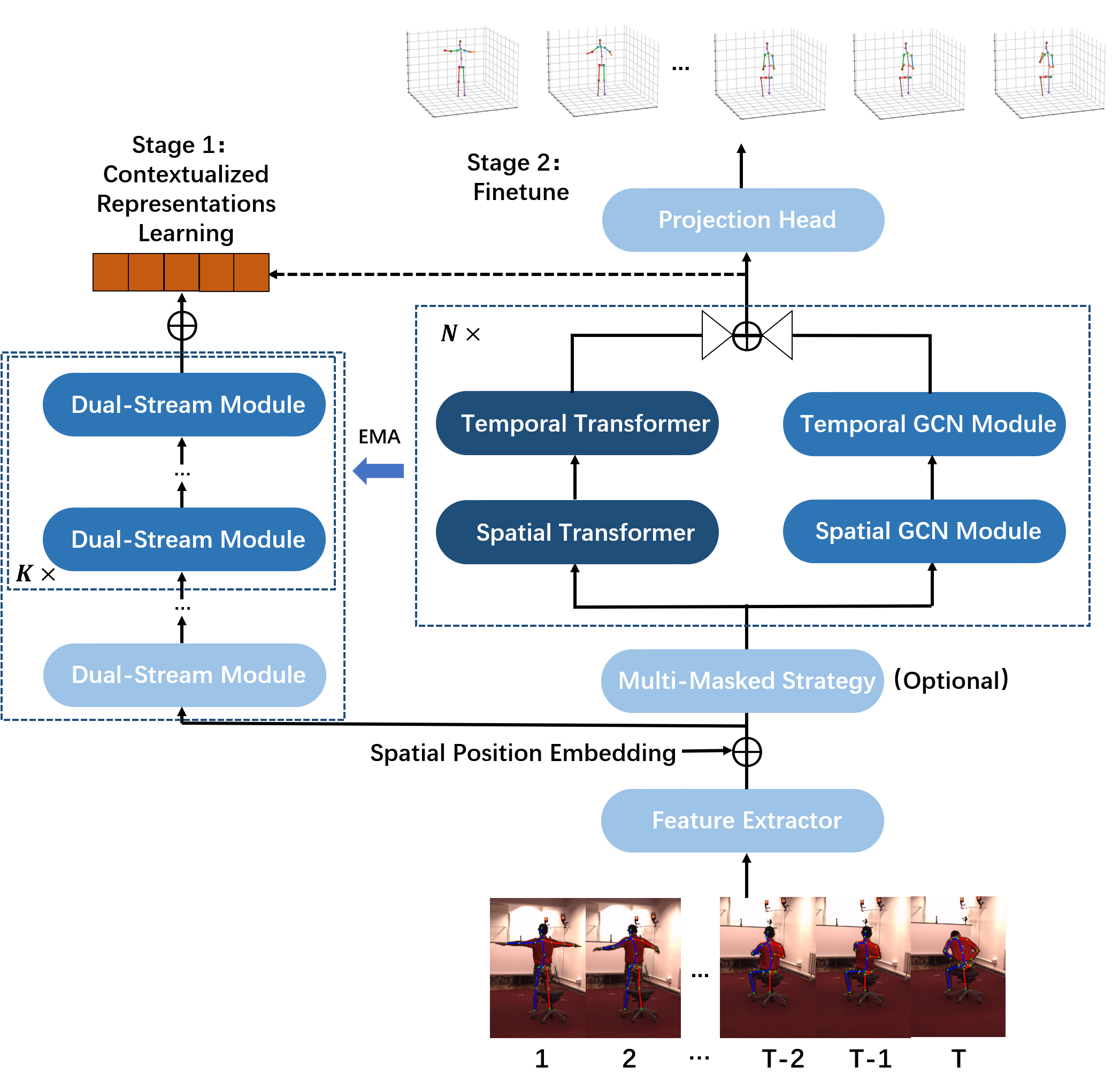}
  \caption{Framework overview. Our method employs a dual-stream architecture with Transformer and GCN to balance local and global dependencies modeling with an adaptive fusion method. The training strategy, divided into pre-training with contextual representation learning and fine-tuning, allows for effectively transfer to robust 3D HPE and even other human motion tasks.}
  \label{fig:fig1}
\end{figure*}

To summarize, although significant achievements have been made through existing research, several challenges remain for monocular 3D HPE methods. (1) 3D HPE tasks require extensive and diverse 3D-labeled datasets to train robust models. However, existing datasets are limited, as most are collected in constrained laboratory environments. (2) 3D HPE for videos necessitates learning 2D-to-3D spatial and temporal correlations, making the complexity of this regression problem notably high. Thus, directly optimizing the model presents a significant challenge. (3) Most state-of-the-art (SOTA) methods for 3D HPE are based on Transformer architectures, which emphasize capturing long-range dependencies. However, 3D HPE tasks inherently focus on both global and local dependencies. To address these challenges, we propose a Transformer-GCN dual-stream model with contextualized representation learning. The overview framework of our method is illustrated in Figure \ref{fig:fig1}.

The model architecture for the training backbone consists of two branches: the Transformer stream and the GCN stream. The GCN stream excels at learning local dependencies for adjacent keypoints and frames, while the Transformer stream captures global spatial and temporal interactions. We employ an adaptive fusion method to combine these two features, generating high-dimensional representations that are both locally and globally balanced.

For the training strategy, we divide the task into a pre-training stage and a fine-tuning stage. In the pre-training stage, we propose a pose estimation pre-training task based on contextualized representation learning to address the issue of limited 3D-labeled datasets for training. After extracting features from 2D pose inputs, we use masked representations to predict latent representations of the full version in a self-distillation setup. This process enables the model to learn spatio-temporal correlations and obtain improved initialization. Unlike \cite{shan2022p, lin2021end, zhu2023motionbert}, we do not predict 2D or 3D poses in the pre-training stage but instead predict contextual representations for motion information. This approach allows us to utilize a large amount of unlabeled 2D pose data, mitigating the need for expensive 3D-labeled data. Additionally, compared to directly predicting 2D poses, our method of contextual feature learning captures advanced high-dimensional representations, allowing the pre-trained model to be directly transferred to the 3D HPE task and even other human motion tasks, such as human action recognition.

In the fine-tuning stage, we fine-tune the pre-trained encoder and predict 3D poses using the Transformer-GCN dual-stream model. Through this two-stage training, the model encoder captures both temporal and spatial dependencies in a globally and locally balanced manner.

In summary, the main contributions of our paper are listed below.

\begin{itemize}
\item{We propose a pre-training method for monocular 3D HPE based on contextualized representation learning, directly addressing the challenge of limited 3D-labeled data. By leveraging unlabeled 2D pose data and employing spatial-temporal modeling, our approach enables the model to learn complex spatial-temporal relationships between postures, significantly enhancing generalization and performance.}
\item{We propose to use the Transformer-GCN dual-stream model, the model can learn both temporal and spatial dependencies in a globally and locally balanced strategy.}
\item{Our model achieves state-of-the-art performance on the Human3.6M and MPI-INF-3DHP benchmarks, with MPJPE scores of 38.0mm and 15.9mm, respectively.}
\end{itemize}

\section{Related Work}

\subsection{3D human pose estimation}

Thanks to advancements in 2D pose estimation \cite{newell2016stacked, sun2019deep, li2021tokenpose, li2022simcc, toshev2014deeppose, luvizon2019human, geng2023human}, two-stage methods (2D-to-3D lifting methods) have become the SOTA in monocular 3D HPE. These methods utilize existing 2D estimation models to obtain 2D poses from images and videos in the first stage, and then use 2D-to-3D lifting models to learn 3D poses from the 2D information in the second stage. By simplifying the data format with intermediate 2D pose inputs, the model can effectively learn from long-range inputs that contain rich temporal and spatial information provided by videos.

Pose estimation for videos can intuitively be regarded as a sequence modeling problem in the temporal dimension, thus sequence models such as RNN, CNN, LSTM, and Transformer are widely used in this context. For example, \cite{hossain2018exploiting} utilizes the temporal information of human pose sequences for 3D pose estimation with LSTM. \cite{bai1803empirical} proposes learning temporal information with TCN and techniques including causal convolution, dilated convolution, and residual connections. Videopose \cite{pavllo20193d} takes the 2D pose sequence from videos as input and uses dilated convolution to extract the temporal information in a semi-supervised paradigm. Li \textit{et al.} \cite{li2022exploiting} identify information redundancy when dealing with long-range pose sequences and propose the Strided Transformer, which combines CNN and Transformer architectures for high efficiency and performance. Based on the Strided Transformer, Shan \textit{et al.} \cite{shan2022p} introduce a self-supervised pre-training method called P-STMO. The pre-training strategy of P-STMO is inspired by MAE \cite{he2022masked}, pre-training the model in a self-reconstruction manner to capture both spatial and temporal dependencies.

Human pose inherently presents a graph structure, where keypoints of the human body can be viewed as vertices and skeletons as edges. The 3D HPE task is structured with rich spatial dependencies, leading many researchers to explore graph-based approaches. Zhao \textit{et al.} \cite{zhao2019semantic} propose Semantic-GCN (SemGCN) to learn edge channel weights through the graph's adjacency matrix. By interleaving SemGConv layers with non-local layers, the model captures both local and global relationships between keypoints. Zou \textit{et al.} \cite{zou2021modulated} introduce an improved GCN with a weight module and an affine module. The weight module performs decoupled feature transformations for different keypoints, while the affine module explores additional joint correlations beyond the defined human skeleton through variations. GraFormer \cite{zhao2021graformer} combines Transformer and Chebyshev GCN \cite{defferrard2016convolutional}, learning long-range dependencies via self-attention mechanisms and local dependencies through GCN. However, research shows that most GCN-only methods are less effective than Transformer-based approaches in extracting global representations.

SOTA 3D HPE methods construct hybrid modules for both temporal and spatial dimensions. Poseformer \cite{zheng20213d} and MixSTE \cite{zhang2022mixste} use pure Transformer blocks to jointly encode temporal and spatial information for pose sequences. STCFormer \cite{tang20233d} sequentially stacks multiple Spatio-Temporal Criss-cross attention blocks and incorporates Structure-enhanced Positional Embedding, which captures local structural relationships through spatio-temporal convolutions and part-aware information to define the body parts each joint belongs to. MotionBERT \cite{zhu2023motionbert} designs a cross spatio-temporal model with Transformer encoders and performs 2D to 3D pre-training on a large amount of 3D-labeled data, achieving excellent results on multiple downstream tasks. RePOSE \cite{sun2024repose} presents a straightforward yet highly effective solution to address the challenges of occlusion. By incorporating spatio-temporal relational depth consistency into the supervision signals, it shifts the focus from absolute depth to relative positioning, improving the robustness and accuracy of pose estimation under occlusions. Although these methods have achieved remarkable accuracy, Transformer-based methods are challenging to train and have inherent issues such as high computational costs. For example, MotionBERT requires extensive data during the pre-training stage.

Other researchers introduce GCN to encode spatial information, leveraging GCN's low computational complexity and ability to learn local dependencies. GLA-GCN \cite{yu2023gla} introduces an adaptive GCN method to learn global representations in the spatial dimension and capture local temporal information with strided modules that hierarchically reduce the receptive scope. DC-GCT \cite{kang2023double} proposes a GCN-based constraint module and a self-attention constraint module to exploit local and global dependencies of the input sequence. Similarly, MotionAGFormer \cite{mehraban2024motionagformer} designs a dual-stream network, using parallel Transformer and GCNFormer streams to jointly perform spatial and temporal modeling, followed by adaptive fusion of these features. Our method synthesizes the influences of the aforementioned approaches.

\begin{figure*}[ht]
  \centering
  \includegraphics[scale=0.32]{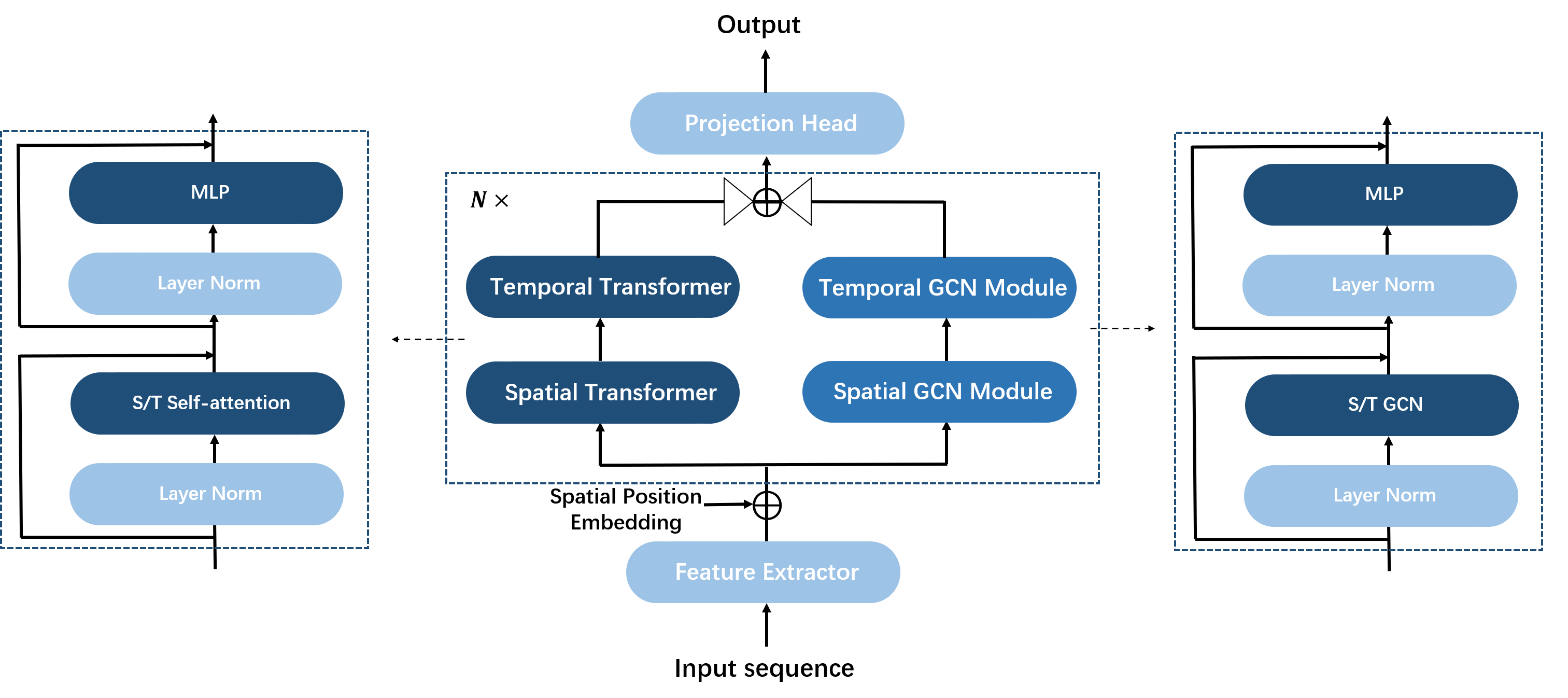}
  \caption{Detailed architecture of Transformer-GCN model. The network consists of two streams, the Transformer stream and the GCN stream. The GCN stream and the Transformer stream are responsible for learning local relations and global interaction respectively, and we use adaptive fusion method to fuse features and generate a new representation that is balanced in spatial-tempoal correlations and local-global modeling.}
  \label{fig:fig2}
\end{figure*}

\subsection{Pre-training technique}

The Transformer has become a mainstream model architecture across various domains, including natural language processing (NLP), computer vision (CV), and speech processing. Its successful adaptations are largely attributed to pre-training techniques. Pre-training involves training models on large-scale datasets, enabling the pre-trained models to be transferred to specific tasks. This technique helps models learn well-trained initializations from extensive datasets, thereby accelerating convergence and improving generalization in downstream tasks. Generally, pre-training techniques can be categorized into self-supervised learning, self-reconstruction learning, and contrastive learning. Our research focuses on self-supervised methods, which leverage the inherent characteristics of inputs for training without manual annotation, thereby enhancing the model's generality.

In NLP, BERT and BERT based methods \cite{devlin2018bert} employs bidirectional Transformer encoders and the Masked Language Modeling (MLM) task to learn contextual and semantic relationships, achieving strong performance across various NLP tasks. Similarly, in CV, Kaiming \textit{et al.} \cite{he2022masked} proposed Masked Autoencoders (MAE), which use a Masked Image Modeling (MIM) task with an asymmetric encoding-decoding strategy to efficiently pre-train models for applications like object detection and semantic segmentation. Recently, Meta introduced Data2vec \cite{baevski2022data2vec} and Data2vec2.0 \cite{baevski2023efficient}, which design a unified multimodal self-supervised learning framework. Data2vec employs a distilled self-supervised architecture, predicting latent representations of the full inputs based on a masked version in a self-distillation setup grounded in the Transformer architecture. Data2vec2.0 builds upon this framework by adopting an efficient strategy that encodes only the masked parts, fills the masked parts with Gaussian noise, and decodes the prediction masks through a simple decoder.

Meanwhile, pre-training techniques have been initially adopted for 3D HPE. P-STMO \cite{shan2022p} introduces self-supervised learning to 3D HPE with the masked pose modeling (MPM) task. During the pre-training stage, P-STMO randomly masks input features in the spatial and temporal dimensions and restores the original 2D pose using Transformer encoders and a simple decoder. Additionally, METRO \cite{lin2021end} and MotionBERT \cite{zhu2023motionbert} propose similar masking pre-training tasks, extending the pre-trained model to human mesh recovery and human action recognition. Although these methods are essentially different from traditional self-supervised learning methods, as they resemble data augmentation techniques requiring extra 3D-labeled data, they have demonstrated significant effectiveness.

Inspired by these pre-training methods, we propose contextualized representations learning methods with the Transformer-GCN model for 3D HPE.

\section{Method}

\subsection{Transformer-GCN dual-stream model}
\label{section:model}

Two-stage monocular 3D HPE algorithms derive 3D poses from 2D pose inputs, tacking this task as a regression problem. Let us denote the input sequence as ${x} \in \mathbb{R} ^ {T \times J \times 3 }$, where ${T}$ and ${J}$ represent the number of frames and joints, respectively. The three-dimensional features represent the 2D coordinates along with the confidence levels of the 2D pose inputs.

As illustrated in Figure \ref{fig:fig2}, the initial step involves mapping the input into a set of high-dimensional features using the feature extraction layer. Specifically, the embedding layer, implemented as MLP, projects the input into ${d}$-dimensional representations, denoted as ${F^{(0)} \in \mathbb{R}^{T \times J \times d}}$. Subsequently, the spatial position embedding ${P^s \in \mathbb{R}^{1 \times J \times d}}$ is added to the feature ${F^{(0)}}$ through broadcasting. Following the approach of \cite{mehraban2024motionagformer}, we omit the use of temporal position embedding since the temporal adjacency matrix in the GCN stream inherently encodes temporal position information. After integrating the positional information, we employ Transformer-GCN dual-stream modules to compute the characteristics of the ${i}$-th layer, represented as ${F^{(i)} \in \mathbb{R}^{T \times J \times d}}$ for ${i=1,2,\ldots,N}$, where ${N}$ denotes the total number of modules. Furthermore, an adaptive fusion method aggregates the features extracted from both the Transformer and GCN streams, formulated as:

\begin{equation}
\label{equation}
F^{(i)}=\alpha_{Tr}^{(i)} \cdot {F_{Tr}}^{(i-1)} + {\alpha_G}^{(i)} \cdot {F_G}^{(i-1)},
\end{equation}
where ${{F_{Tr}}^{(i)}}$ and ${{F_G}^{(i)}}$ are the intermediate representations of the ${i}$-th layer in the Transformer and GCN streams, respectively. The adaptive fusion weights, ${\alpha_{Tr}^{(i)}}$ and ${\alpha_{G}^{(i)}}$, are computed as follows:

\begin{small}
\begin{equation}
{ {\alpha_{Tr}^{(i)},\alpha_{G}^{(i)}} } =  softmax\left( {W \cdot Concat\left( {F_{Tr}^{({i - 1})}, F_{G}^{({i - 1})}} \right)} \right),
\end{equation}
\end{small}

\noindent where ${softmax}$ denotes the activation function, ${Concat}$ is the concatenation function, and ${W}$ represents the learnable weight for the linear layer. Both the Transformer and GCN streams consist of spatial and temporal modules. The spatial module treats each human keypoint as a distinct feature, effectively capturing the spatial correlation among different joints. Conversely, the temporal module treats each frame as a distinct feature, capturing the inter-frame relationships. Finally, a projection head maps the features into 3D space, yielding the 3D pose output ${{\Tilde{X}} \in \mathbb{R}^{T \times J \times 3}}$.

\subsubsection{Transformer stream}

The model architecture within the Transformer stream is primarily constructed using the Transformer encoder. The spatial Transformer module and the temporal Transformer module employ spatial and temporal self-attention mechanisms, respectively, to capture spatial and temporal correlations. For instance, the spatial Transformer module utilizes the multi-head self-attention mechanism to learn the interactions of posture information across multiple high-dimensional spaces from a spatial perspective, formalized as follows:

\begin{small}
\begin{equation}
M_-SA \left( {Q_{S}, K_{S}, V_{S}} \right) =  Concat\left( {{\text{head}}_{1}, \ldots, {\text{head}}_{h}} \right) W_{TrS},
\end{equation}
\end{small}

\noindent where ${Q_{S}}$, ${K_{S}}$, and ${V_{S}}$ denote the query, key, and value in the spatial module. ${W_{TrS}}$ is the learnable projection matrix, and ${{\text{head}}_{k}}, (k=1,2,...,h)$ represents the feature of the ${k}$-th attention head, which can be calculated as:

\begin{equation}
{\text{head}}_{k} = Softmax\left( \frac{Q{S}^{(k)} \left( K_{S}^{(k)} \right)^{T}}{\sqrt{d}} \right) V_{S}^{(k)},
\end{equation}
where ${Q_{S}^{(k)}}$, ${K_{S}^{(k)}}$, and ${V_{S}^{(k)}}$ are defined as:

\begin{equation}
\begin{split}
Q_{S}^{(k)} = F_{TrS} \cdot W_{TrS}^{({Q,k})}, \\
K_{S}^{(k)} = F_{TrS} \cdot W_{TrS}^{({K,k})}, \\
V_{S}^{(k)} = F_{TrS} \cdot W_{TrS}^{({V,k})},
\end{split}
\end{equation}
where ${F_{TrS} \in \mathbb{R}^{BT \times J \times d}}$ represents the spatial module's output in the Transformer stream, with ${B}$ being the batch size. ${W_{TrS}^{({Q,k})}}$, ${W_{TrS}^{({K,k})}}$, and ${W_{TrS}^{({V,k})}}$ denote the learnable projection matrices for query, key, and value, respectively. Similar to the Transformer encoder, our process incorporates feed-forward layers and shortcut techniques, expressed as:

\begin{equation}
\begin{split}
F_{TrS} = F_{TrS} + M_-SA\left( {LN\left( F_{TrS} \right)} \right), \\
F_{TrS} = F_{TrS} + MLP\left( {LN\left( F_{TrS} \right)} \right),
\end{split}
\end{equation}
where ${LN(\cdot)}$ denotes layer normalization. For the temporal module, the overall process is analogous, with the primary difference being that the temporal intermediate feature is defined as ${F_{TrT} \in \mathbb{R}^{BJ \times T \times d}}$.

\subsubsection{GCN stream}

Unlike the Transformer stream, which focuses on capturing global dependencies, the GCN stream emphasizes spatial and temporal interactions within a local context. By combining these two streams, the model balances local and global interactions, integrating information more effectively. Similar to the Transformer stream, we compute information for both spatial and temporal dimensions. The GCN stream's architecture is based on the graph convolution network, which is divided into the spatial GCN module and the temporal GCN module. For instance, the spatial GCN module emphasizes the interaction between adjacent joints \cite{luo2022learning, zhao2019semantic}, calculated as:

\begin{scriptsize}
\begin{equation}
F_{GS} = ReLU\left( {F_{GS} + BN\left( {\Tilde{D}}_{S}^{- \frac{1}{2}}{\Tilde{A}}_{S}{\Tilde{D}}_{S}^{- \frac{1}{2}}F_{GS}W_{1} + F_{GS}W_{2} \right)} \right),
\end{equation}
\end{scriptsize}

\noindent where $ReLU(\cdot)$ is the activation function and $BN(\cdot)$ denotes batch normalization. ${W_{1}}$ and ${W_{2}}$ are the learnable weights, and ${F_{GS} \in \mathbb{R}^{BT \times J \times d}}$ represents the feature in the spatial GCN module. Here, ${\Tilde{A}}_{S} = {A}_{S} + I$, where ${\Tilde{A}}_{S}$ is the adjacency matrix with self-connections, ${A}_{S}$ is the adjacency matrix for the spatial GCN module, and ${I}$ is the identity matrix. ${\Tilde{D}}_{S}$ is the degree matrix of ${\Tilde{A}}_{S}$. Finally, we calculate the spatial feature as:

\begin{equation}
F_{GS} = F_{GS} + MLP\left( LN \left( F_{GS} \right) \right),
\end{equation}

The temporal GCN module operates similarly to the spatial module, with two key differences. First, the temporal feature is defined as ${F_{GT} \in \mathbb{R}^{BJ \times T \times d}}$. Second, the adjacency matrix differs. The adjacency matrix in the spatial module, ${A_{S}}$, is determined by the distribution of human joints, representing the topological relationships between them. For example, the right leg is connected to the right knee, so their corresponding position in the matrix is ${A_{S}(3, 2) = 1}$. Unlike \cite{mehraban2024motionagformer}, we define the adjacency matrix in the temporal module based on the sequential relationship between frames to reduce training cost. Therefore, all elements on the sub-diagonal of ${A_{T}}$ are 1, while the rest are 0.

\subsubsection{Objective}

After obtaining the predicted 3D pose ${{\Tilde{X}}}$, we compare it with the ground truth 3D pose ${X}$ to calculate the losses for model training. We employ three loss functions: the mean per joint position error (MPJPE), the normalized mean per joint position error (N-MPJPE), and the mean per joint velocity error.

MPJPE, the most widely used metric for 3D HPE, directly measures the average position error between the prediction and the ground truth using Euclidean distance. It is defined as:

\begin{equation}
L_{MPJPE}\left( {\Tilde{X}, X} \right) = \frac{1}{TJ} \sum_{t = 1}^{T} \sum_{j = 1}^{J} \left| \middle| {\Tilde{X}{t,j}} - X{t,j} \middle| \right|_{2},
\end{equation}

Since MPJPE averages the error for all joints equally, ignoring the differences between human keypoints, we also use N-MPJPE, a variant of MPJPE that calculates the normalized error to eliminate the influence of scale factors. It is formulated as:

\begin{footnotesize}
\begin{equation}
L_{N-MPJPE}\left( {\Tilde{X}, X} \right) = \frac{\Tilde{X} \cdot X}{\Tilde{X} \cdot \Tilde{X}} \cdot \frac{1}{TJ} \sum_{t = 1}^{T} \sum_{j = 1}^{J} \left| \middle| {\Tilde{X}}{t,j} - X{t,j} \middle| \right|_{2},
\end{equation}
\end{footnotesize}

Given that the model can output multiple frames for the 3D pose, we propose using the mean per joint velocity error to measure the velocity error between the prediction and the annotation. This loss is defined as:

\begin{equation}
L_{Vel}\left( {\Tilde{X}, X} \right) = \frac{1}{\left( {T - 1} \right)J} \sum_{t = 2}^{T} \sum_{j = 1}^{J} \left| \middle| {\Tilde{V}}{t,j} - V{t,j} \middle| \right|_{2},
\end{equation}
where the per-frame per-joint velocity of predictions ${\Tilde{V}{t,j}}$ and the per-frame per-joint velocity of ground truth ${V{t,j}}$ are calculated as:

\begin{equation}
{\Tilde{V}}{t,j} = {\Tilde{X}}{t,j} - {\Tilde{X}}{t - 1,j}, \quad V{t,j} = X_{t,j} - X_{t - 1,j},
\end{equation}

In summary, the total loss during the fine-tuning stage is defined as:

\begin{equation}
L_{Finetune} = L_{MPJPE} + \lambda_{1} L_{N-MPJPE} + \lambda_{2} L_{Vel},
\end{equation}
where ${\lambda_{1}}$ and ${\lambda_{2}}$ are hyperparameters.

\subsection{Contextualized representation learning}

\begin{figure}[ht]
  \centering
  \includegraphics[scale=0.32]{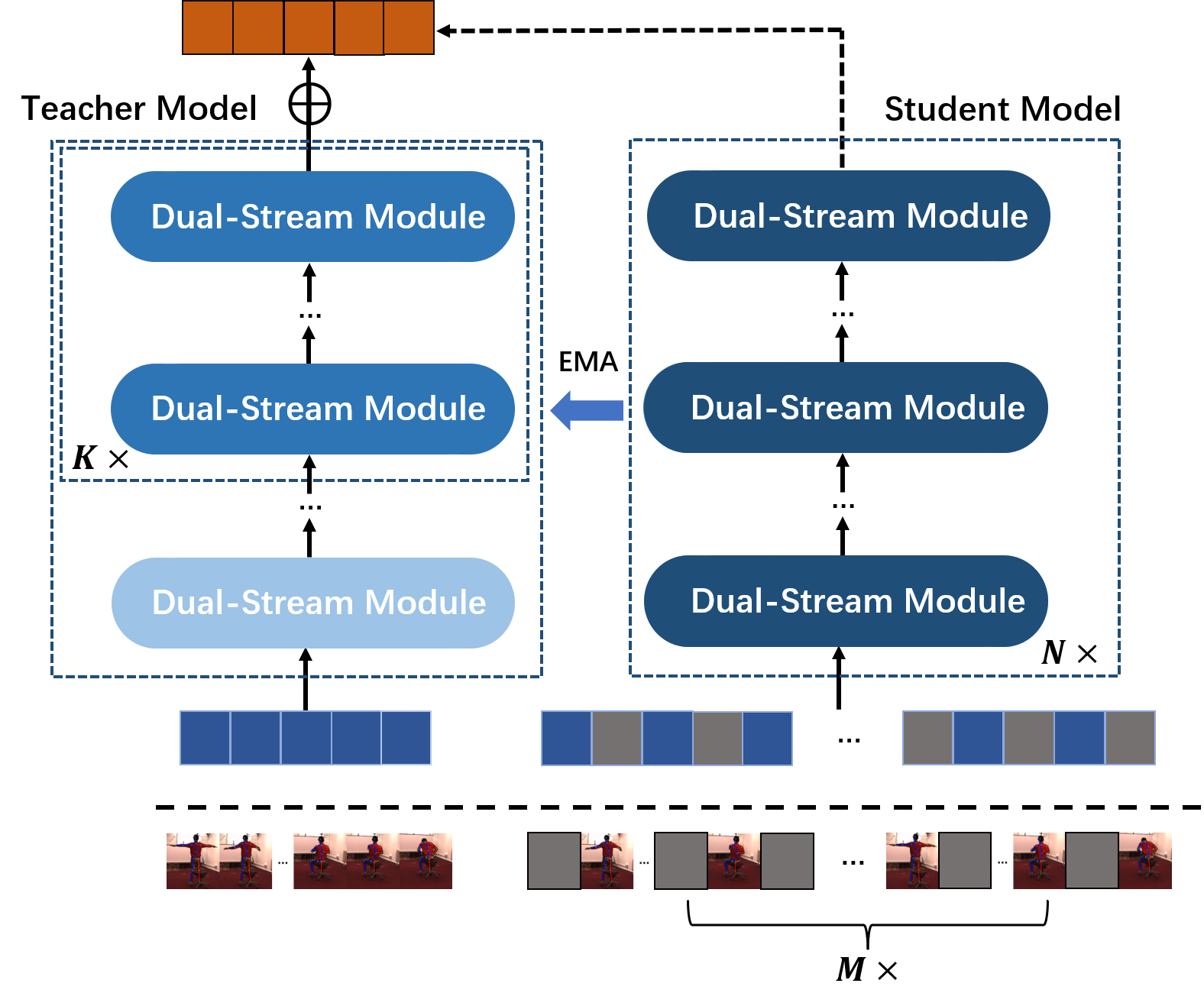}
  \caption{Framework of our pre-training task based on the contextualized representations learning. After extracting features from 2D pose input by the feature extractor layer, we apply our masking strategy and use masked representations to predict latent representations of the full version in a self-distillation setup. Through this process, the model can learn the spatio-temporal correlations and obtain great initialization.}
  \label{fig:fig3}
\end{figure}

After constructed the Transformer-GCN two-stream model in section ~\ref{section:model}, we propose a pose estimation pre-training task to train our model based on the contextualized representations learning. As shown in Figure \ref{fig:fig3}.

Inspired by \cite{shan2022p, baevski2022data2vec, he2022masked, baevski2023efficient}, our pre-training task is primarily based on contextualized representations learning and self-distillation. Specifically, we first utilize the feature extractor layer to extract high-dimensional features from the inputs. Then, we apply a masking strategy to these features, referring to following subsection. To construct the training target, we employ the weighted exponential moving average (EMA) strategy to build the teacher model. The teacher model encodes the unmasked training sample features to obtain the complete features, which serve as the learning target. Finally, the masked learned features are used to predict the training target provided by the teacher model, thereby reconstructing the contextualized representation. In summary, the student model encodes the masked features of pose inputs, while the teacher model encodes the unmasked version of these inputs to construct the training target. The parameters of the teacher model are derived as the EMA result of the student model.

\subsubsection{Masking strategy}
\label{section:masking}

After embedding the inputs into high-dimensional features, we apply a masking strategy to the feature ${F^{(0)}}$. Initially, we receive pose features with batch size ${B}$ and sequence length ${L}$. We then construct a mask matrix ${Mask}$ with a mask probability ${P_{mask}}$, having dimensions ${B \times L}$, ensuring that each row has a mask ratio of ${P_{mask}}$. This mask matrix is then applied to the input features. Instead of using learnable parameters to represent masked parts, we utilize random Gaussian noise \cite{shan2022p, he2022masked}. Gaussian noise is more efficient compared to learnable masked tokens, as it provides regularization and enhances input diversity during training, thereby improving model performance.

To further enhance the pre-training task and provide richer input representations, we propose a multi-masking strategy inspired by \cite{baevski2023efficient}. We consider ${M}$ different masked versions of the input representations and use them to predict the same target feature, calculating the loss of multiple masked features relative to the same target feature. Specifically, we generate ${M}$ mask matrices to construct the total mask matrix ${P_{masks}}$ of size ${BM \times L}$. We then replicate the input features in the student model ${M}$ times to a size of ${BM \times L \times C}$, allowing the mask matrices to be applied to the input representations. Meanwhile, we do not replicate the features in the teacher model but reuse the teacher representations for the training samples.

Our multi-masking strategy not only enriches the input representations for the pre-training stage but also prevents redundant calculations and reduces training costs. By adjusting the batch size ${B}$ as the number of masking strategies ${M}$ increases, we can ensure that the increase in training cost remains negligible.

\subsubsection{Training Target}
\label{section:target}

To construct the training target, we use the teacher model to encode the unmasked features and generate contextualized representations. This design enables the model to learn high-dimensional correlations from unlabeled data during the pre-training stage, thereby providing rich feature representations for human motion tasks such as 3D HPE. We specifically discuss the update mechanism of the teacher model, the training objective, and the loss function below.

For the update mechanism of the teacher model, it is designed to provide contextualized representations as training targets for the student model. Unlike the student model, the parameters of the teacher model do not participate in the training process but are updated using an exponential moving average (EMA) method. The EMA method updates the parameters of the teacher model by taking the weighted average of the parameters of the student model and the historical parameters of the teacher model. This smooth updating process enhances the stability and generalization during training. Specifically, let the parameters of the student model be ${\theta}$ and the parameters of the teacher model be ${\delta}$. The update rule is defined as ${\delta \leftarrow \tau \cdot \delta + \left( 1 - \tau \right) \cdot \theta}$, where ${\tau}$ denotes the exponential weight that follows a linear increase strategy. If ${\tau}$ is closer to 0, the teacher model closely follows the student model. Conversely, if ${\tau}$ is closer to 1, the teacher model remains closer to its historical version. As training progresses, the influence of the student model on the teacher model decreases, allowing the teacher model's parameters to be updated smoothly and gain stability.

Regarding the training objective, unlike \cite{shan2022p, he2022masked, zhu2023motionbert}, we do not directly use the final output of the teacher model as the training target. Instead, we base our objective on the output of the Top-K network modules in the teacher model. According to Equation \ref{equation:equ14}, we obtain the output of the Top-K modules as ${{ \Tilde{F}^{(i)}, (i=N-K+1, \ldots, N) }}$. We also normalize the representations to prevent training collapse and avoid domination by modules with high norms. The overall training objective ${\Tilde{F}}$ is defined as:

\begin{equation}
\Tilde{F} = LN\left( \frac{1}{K} \sum_{i = N-K+1}^{N} LN\left( \Tilde{F}^{(i)} \right) \right),
\label{equation:equ14}
\end{equation}
where ${LN(\cdot)}$ denotes layer normalization.

For the training loss, let the masked-version prediction output by the student model be ${F}$ and the learning objective output by the teacher model be ${\Tilde{F}}$. The total loss for the pre-training stage is the L2 loss between ${F}$ and ${\Tilde{F}}$.

\section{Experiments}

\subsection{Dataset and evaluation metric}

For both pre-training and fine-tuning stages, our method was trained and evaluated on two popular 3D HPE datasets: Human3.6M \cite{ionescu2013human3} and MPI-INF-3DHP \cite{mehta2017monocular}.

\textbf{Human3.6M} is one of the largest motion capture datasets, focusing on indoor scenes. It comprises a total of 3.6 million frames of video data captured by four 50-frame high-resolution scanning cameras. The dataset features up to 17 poses of daily activities performed by 11 actors. We follow the standard data split, using 5 subjects for training (S1, S5, S6, S7, S8) and 2 subjects for testing (S9, S11). For evaluation metrics, we adhere to previous standards \cite{pavllo20193d, zheng20213d, tang20233d, zhao2023poseformerv2} and employ two protocols to comprehensively assess the model's accuracy and robustness. Protocol 1 uses MPJPE, which directly measures the mean Euclidean distance between the predictions and the ground truth in millimeters. Protocol 2 employs procrustes MPJPE (P-MPJPE), which applies rigid transformations such as rotation, translation, and alignment to the predictions to align them with the real poses. Thus, P-MPJPE provides a more comprehensive evaluation of the model's performance in real-world scenarios.

\textbf{MPI-INF-3DHP} is a challenging dataset that includes both restricted indoor and complex outdoor scenes. It features 8 actors performing 8 different actions in various scenes, recorded with 14 cameras, resulting in over 1.3 million frames of data available for training and testing. We follow the standard approach by using 6 subjects for training and 2 subjects for testing. MPJPE is used as the primary evaluation metric for MPI-INF-3DHP.

\subsection{Implementation Details}

The experiments were conducted on a system running Ubuntu 20.04 with CUDA version 12.2. We utilized Python 3.8 as the programming language and PyTorch 2.0 as the primary deep learning framework. The hardware setup included two NVIDIA RTX 4090 GPUs, providing substantial computational power for training and testing.

Before comparing our model with other existing SOTA methods on the two datasets, we employed Yolo \cite{wu2021application} as the object detector and Stacked Hourglass \cite{newell2016stacked} as the 2D pose detector to obtain 2D pose sequences and confidences for predictions. Additionally, we used the 2D ground truth provided by the Human3.6M dataset for experiments, with confidence set to 1.

For our base model's experimental parameters, the temporal length ${T}$ was set to 243 with ${J=17}$ joints. The dual-stream network comprised ${N=16}$ layers, and the feature dimension ${d}$ was set to 128. Regarding the attention mechanism settings, the number of attention heads ${h}$ was set to 8, and the hidden layer feature dimension ratio ${r}$ was 4. The parameters for the large model mirrored those of the base model, except that the feature dimension ${d}$ was set to 256. During the pre-training stage, a masking probability ${P}$ of 0.8 was used, with a multi-mask strategy involving ${M=3}$ masked versions. The pre-training learning objective included 8 layers for ${K}$. For the loss functions, the hyperparameter for N-MPJPE ${\lambda_1}$ was set to 0.5, and the hyperparameter for velocity loss ${\lambda_2}$ was set to 20.

\subsection{Quantitative results}

\begin{table*}[!t]
\caption{Quantitative Results of 3D Human Pose Estimation on Human3.6M. 
(Top) Protocol 1 Using Detected 2D Pose Sequences. (Middle) Protocol 2 Using Detected 2D Pose Sequences. (Bottom) MPJPE (mm) Using Ground Truth 2D Pose Sequences. The Best and Second-Best Results are Highlighted in Bold and Underlined Formats. 
}
\center
\setlength{\tabcolsep}{5pt}
\resizebox{1\linewidth}{!}{
\begin{tabular}{l | c c c c c c c c c c c c c c c c}

\hline 
Method & Dire. & Disc. & Eat & Greet & Phone & Photo & Pose & Purch. & Sit & SitD & Smoke & Wait & WalkD & Walk & WalkT & Avg. \\
\hline 
VideoPose3D~\cite{pavllo20193d} CVPR'19 & 45.2 & 46.7 & 43.3 & 45.6 & 48.1 & 55.1 & 44.6 & 44.3 & 57.3 & 65.8 & 47.1 & 44.0 & 49.0 & 32.8 & 33.9 & 46.8 \\

SRNet~\cite{zeng2020srnet} ECCV'20 & 46.6 & 47.1 & 43.9 & 41.6 & 45.8 & 49.6 & 46.5 & 40.0 & 53.4 & 61.1 & 46.1 & 42.6 & 43.1 & 31.5 & 32.6 & 44.8 \\

PoseFormer~\cite{zheng20213d} ICCV'21 & 41.5 & 44.8 & 39.8 & 42.5 & 46.5 & 51.6 & 42.1 & 42.0 & 53.3 & 60.7 & 45.5 & 43.3 & 46.1 & 31.8 & 32.2 & 44.3 \\

Anatomy~\cite{chen2021anatomy} TCSVT'21 & 41.4 & 43.5 & 40.1 & 42.9 & 46.6 & 51.9 & 41.7 & 42.3 & 53.9 & 60.2 & 45.4 & 41.7 & 46.0 & 31.5 & 32.7 & 44.1 \\

Lifting~\cite{li2022exploiting} TMM'22 & 39.9 & 43.4 & 40.0 & 40.9 & 46.4 & 50.6 & 42.1 & 39.8 & 55.8 & 61.6 & 44.9 & 43.3 & 44.9 & 29.9 & 30.3 & 43.6 \\

PAA~\cite{xue2022boosting} TIP'22 & 39.9 & 42.7 & 40.3 & 42.3 & 45.0 & 52.8 & 40.4 & 39.3 & 56.9 & 61.2 & 44.1 & 41.3 & 42.8 & 28.4 & 29.3 & 43.1 \\

MixSTE~\cite{zhang2022mixste} CVPR'22 & 36.7 & 39.0 & 36.5 & 39.4 & \underline{40.2} & \textbf{44.9} & 39.8 & 36.9 & 47.9 & 54.8 & \textbf{39.6} & 37.8 & 39.3 & 29.7 & 30.6 & 39.8 \\

P-STMO~\cite{shan2022p} ECCV'22 & 38.4 & 42.1 & 39.8 & 40.2 & 45.2 & 48.9 & 40.4 & 38.3 & 53.8 & 57.3 & 43.9 & 41.6 & 42.2 & 29.3 & 29.3 & 42.1 \\

GLA-GCN~\cite{yu2023gla} ICCV'23 & 41.3 & 44.3 & 40.8 & 41.8 & 45.9 & 54.1 & 42.1 & 41.5 & 57.8 & 62.9 & 45.0 & 42.8 & 45.9 & 29.4 & 29.9 & 44.4 \\

STCFormer~\cite{tang20233d} CVPR'23 & 38.4 & 41.2 & 36.8 & 38.0 & 42.7 & 50.5 & 38.7 & 38.2 & 52.5 & 56.8 & 41.8 & 38.4 & 40.2 & 26.2 & 27.7 & 40.5 \\

MotionBERT~\cite{zhu2023motionbert} ICCV'23 & \underline{36.3} & \underline{38.7} & 38.6 & 33.6 & 42.1 & 50.1 & \textbf{36.2} & 35.7 & \underline{50.1} & 56.6 & 41.3 & 37.4 & 37.7 & 26.6 & \underline{26.5} & 39.2 \\

MotionAGFormer~\cite{mehraban2024motionagformer}  WACV'24 & - & - & - & - & - & - & - & - & - & - & - & - & - & - & - & 38.4 \\
\hline
Ours-B & \textbf{36.0} & 39.3 & \underline{36.7} & \underline{33.4} & 40.5 & \underline{48.8} & \underline{36.4} & \underline{35.3} & 50.5 & \underline{52.0} & 40.9 & \underline{36.8} & \textbf{36.3} & \underline{26.6}  & \textbf{26.4} & \underline{38.4} \\
Ours-L & 36.5 & \textbf{38.1} & \textbf{36.0} & \textbf{32.5} & \textbf{39.9} & 49.3 & 36.7 & \textbf{34.7} & \textbf{49.0} & \textbf{50.7} & \underline{40.3} & \textbf{36.7} & \underline{36.9} & \textbf{26.1} & 26.7 & \textbf{38.0} \\
\hline
\hline 
VideoPose3D~\cite{pavllo20193d} CVPR'19 & 34.1 & 36.1 & 34.4 & 37.2 & 36.4 & 42.2 & 34.4 & 33.6 & 45.0 & 52.5 & 37.4 & 33.8 & 37.8 & 25.6 & 27.3 & 36.5 \\

PoseFormer~\cite{zheng20213d} ICCV'21 & 32.5 & 34.8 & 32.6 & 34.6 & 35.3 & 39.5 & 32.1 & 32.0 & 42.8 & 48.5 & 34.8 & 32.4 & 35.3 & 24.5 & 26.0 & 34.6 \\

Anatomy~\cite{chen2021anatomy} TCSVT'21 & 32.6 & 35.1 & 32.8 & 35.4 & 36.3 & 40.4 & 32.4 & 32.3 & 42.7 & 49.0 & 36.8 & 32.4 & 36.0 & 24.9 & 26.5 & 35.0 \\

Lifting~\cite{li2022exploiting} TMM'22 & 32.0 & 35.6 & 32.6 & 35.1 & 35.8 & 40.9 & 33.2 & 31.2 & 44.7 & 48.9 & 36.7 & 33.9 & 35.2 & 23.7 & 25.1 & 34.9 \\

PAA~\cite{xue2022boosting} TIP'22 & 31.2 & 34.1 & 31.9 & 33.8 & 33.9 & 39.5 & 31.6 & 30.0 & 45.4 & 48.1 & 35.0 & 31.1 & 33.5 & 22.4 & 23.6 & 33.7 \\

MixSTE~\cite{zhang2022mixste} CVPR'22 & 30.8 & 33.1 & \textbf{30.3} & 31.8 & \textbf{33.1} & 39.1 & 31.1 & 30.5 & 42.5 & \underline{44.5} & \textbf{34.0} & 30.8 & 32.7 & \underline{22.1} & 22.9 & 32.6 \\

P-STMO~\cite{shan2022p} ECCV'22 & 31.3 & 35.2 & 32.9 & 33.9 & 35.4 & 39.3 & 32.5 & 31.5 & 44.6 & 48.2 & 36.3 & 32.9 & 34.4 & 23.8 & 23.9 & 34.4 \\

STCFormer~\cite{tang20233d} CVPR'23 & 29.5 & 33.2 & 30.6 & 31.0 & 33.0 & 38.0 & 30.4 & 29.4 & 41.8 & 45.2 & 33.6 & 29.5 & 31.6 & 21.3 & 22.6 & 32.0 \\

GLA-GCN~\cite{yu2023gla} ICCV'23 & 32.4 & 35.3 & 32.6 & 34.2 & 35.0 & 42.1 & 32.1 & 31.9 & 45.5 & 49.5 & 36.1 & 32.4 & 35.6 & 23.5 & 24.7 & 34.8 \\

MotionAGFormer~\cite{mehraban2024motionagformer}  WACV'24 & - & - & - & - & - & - & - & - & - & - & - & - & - & - & - & 32.5 \\
\hline
Ours-B & \textbf{30.1} & \underline{32.5} & 31.4 & \underline{28.0} & 33.7 & \textbf{38.1} & \textbf{29.6} & \underline{30.0} & \underline{42.3} & 45.4 & 35.1 & \underline{29.9} & \textbf{30.7} & 22.4 & \underline{22.8} & \underline{32.1} \\
Ours-L & \underline{30.3} & \textbf{32.0} & \underline{30.9} & \textbf{27.5} & \underline{33.2} & \underline{38.9} & \underline{30.1} & \textbf{29.5} & \textbf{40.7} & \textbf{44.2} & \underline{34.6} & \textbf{29.8} & \underline{31.6} & \textbf{22.2} & \textbf{22.8} & \textbf{31.9} \\
\hline
\hline
PoseFormer~\cite{zheng20213d} ICCV'21 & 30.0 & 33.6 & 29.9 & 31.0 & 30.2 & 33.3 & 34.8 & 31.4 & 37.8 & 38.6 & 31.7 & 31.5 & 29.0 & 23.3 & 23.1 & 31.3 \\

Lifting~\cite{li2022exploiting} TMM'22 & 27.9 & 29.9 & 26.8 & 27.8 & 28.6 & 32.8 & 31.1 & 26.7 & 36.5 & 35.5 & 30.0 & 29.8 & 27.5 & 19.6 & 19.7 & 28.6 \\

PAA~\cite{xue2022boosting} TIP'22 & 25.8 & 25.2 & 23.3 & 23.5 & 24.0 & 27.4 & 27.9 & 24.4 & 29.3 & 30.1 & 24.9 & 24.1 & 23.3 & 18.6 & 17.9 & 24.7 \\

MixSTE~\cite{zhang2022mixste} CVPR'22 & 21.6 & 22.0 & 20.4 & 21.0 & 20.8 & 24.3 & 24.7 & 21.9 & \underline{26.9} & 24.9 & \underline{21.2} & 21.5 & 20.8 & \textbf{14.7} & 15.6 & 21.6 \\

P-STMO~\cite{shan2022p} ECCV'22 & 28.5 & 30.1 & 28.6 & 27.9  & 29.8 & 33.2 & 31.3 & 27.8 & 36.0 & 37.4 & 29.7 & 29.5 & 28.1 & 21.0 & 21.0 & 29.3 \\

GLA-GCN~\cite{yu2023gla} ICCV'23 & \underline{20.2} & \underline{21.9} & 21.7 & 19.9 & 21.6 & 24.7 & 22.5 & \underline{20.8} & 28.6 & 33.1 & 22.7 & \underline{20.6} & 20.3 & 15.9 & 16.2 & 22.0 \\

STCFormer~\cite{tang20233d} CVPR'23 & 20.8 & 21.8 & 20.0 & 20.6 & 23.4 & 25.0 & 23.6 & 19.3 & 27.8 & 26.1 & 21.6 & 20.6 & 19.5 & 14.3 & 15.1 & 21.3 \\

MotionAGFormer~\cite{mehraban2024motionagformer}  WACV'24 & - & - & - & - & - & - & - & - & - & - & - & - & - & - & - & \underline{19.4} \\
\hline
Ours-B & \textbf{19.3} & 22.6 & \underline{20.7} & \underline{20.2} & \underline{19.7} & \textbf{20.4} & \underline{21.6} & 24.0 & \textbf{25.2} & \underline{20.8} & 24.5 & \textbf{18.1} & \textbf{10.4} & \underline{15.7} & \underline{16.1} & 20.6 \\

Ours-L & 20.6 & \textbf{19.1} & \textbf{19.5} & \textbf{18.9} & \textbf{17.9} & \underline{20.8} & \textbf{18.7} & \textbf{18.4} & 27.6 & \textbf{20.4} & \textbf{20.7} & 21.0 & \underline{13.6} & 15.9 & \textbf{13.6} & \textbf{19.3} \\ 
\hline

\end{tabular}}
\vspace{0.1cm}
\label{tab:table1}
\end{table*}

We compare our method with other existing methods on the Human3.6M dataset, considering only those trained without additional data for a fair comparison. The results are presented in Table \ref{tab:table1}.

In Protocol 1, our large model achieves a SOTA result with the MPJPE of 38.0 mm and our base model achieves the MPJPE of 38.4 mm. Our experiment provides detailed results for each action, achieving the best performance in most actions, especially those with rich information such as eating, greeting, and walking together. These results demonstrate the accuracy and robustness of our model.

For Protocol 2, our model surpasses nearly all methods, achieving the MPJPE of 31.9 mm with the best results in most actions. Additionally, by directly using the 2D ground truth from the Human3.6M dataset to train the model, our model achieves the MPJPE of 19.3 mm. We observe that the more accurate the provided 2D pose data, the better the performance our model achieves.

\begin{table}[htbp]
\center
\caption{Quantitative Results on MPI-INF-3DHP.}
\begin{tabular}{l | ccc}
\hline
Model            & MPJPE $\downarrow$ & PCK $\uparrow$ & AUC $\uparrow$ \\
\hline
Poseformer~\cite{zheng20213d}                      & 77.4 & 88.6 & 56.4 \\
P-STMO~\cite{shan2022p}                            & 32.2 & 97.9 & 75.8 \\
MixSTE~\cite{zhang2022mixste}                      & 54.9 & 94.4 & 66.5 \\
GLA-GCN~\cite{yu2023gla}                           & 27.7 & 98.5 & 79.1 \\
STCFormer~\cite{tang20233d}                        & 23.1 & 98.7 & 83.9 \\ 
MotionAGFormer-B~\cite{mehraban2024motionagformer} & 18.2 & 98.3 & 84.2 \\
MotionAGFormer-L~\cite{mehraban2024motionagformer} & 16.2 & 98.2 & 85.3 \\
\hline
Ours-B                                             & 16.5 & 98.7 & 86.4 \\
Ours-L                                             & 15.9 & 98.5 & 87.0 \\
\hline
\end{tabular}
\vspace{0.1cm}
\label{tab:table2}
\end{table}

The MPI-INF-3DHP dataset presents a greater challenge as it includes both indoor and outdoor scenes, and the amount of training and testing data is not as extensive as that of Human3.6M. As shown in Table \ref{tab:table2}, our basic model (Ours-B) achieves an MPJPE of 16.5 mm, surpassing MotionAGFormer-B by approximately 1.7 mm. Our large model reaches an MPJPE of 15.9 mm with 0.3 mm lower than MotionAGFormer-L's result. Both accuracy and PCK also undergo considerable improvement compared to the baseline model, as accuracy improved from 85. 3\% to 87. 0\% and PCK increased to 98. 7\%. These results highlight the robustness of our model for complex scene shown in this dataset.

\begin{table}[htbp]
\center
\caption{Complexity Analysis of Our Model.}
\begin{tabular}{@{}l | cccc@{}}
\hline
Model                                              & Param(M) $\downarrow$ & MACs(G) $\downarrow$ & FPS $\uparrow$  & MPJPE $\downarrow$ \\
\hline
Poseformer~\cite{zheng20213d}                      & 9.5      & 0.8     & 327   & 44.3  \\
P-STMO~\cite{shan2022p}                            & 6.7      & 1.0     & 148   & 42.8  \\
MixSTE~\cite{zhang2022mixste}                      & 33.6     & 147.6   & 9811  & 40.9  \\
GLA-GCN~\cite{yu2023gla}                           & 1.3      & 1.6     & 79    & 44.4  \\
MotionBERT~\cite{zhu2023motionbert}                & 42.5     & 174.7   & 8021  & 39.2  \\
MotionAGFormer-B~\cite{mehraban2024motionagformer} & 11.7     & 64.8    & 6511  & 38.4  \\
MotionAGFormer-L~\cite{mehraban2024motionagformer} & 19.0     & 105.1   & 4026  & 38.4  \\
\hline
Ours-B                                             & 11.7     & 48.0    & 6588  & 38.4  \\
Ours-L                                             & 46.4     & 91.8    & 4558  & 38.0  \\
\hline
\end{tabular}
\vspace{0.1cm}
\label{tab:table3}
\end{table}

As detailed in Table \ref{tab:table3}, our model has 11.7 M parameters and 48.0G MACs. Compared to other methods, our model maintains a suitable size and computational complexity while achieving SOTA performance. Specifically, our method outperforms smaller models like P-STMO \cite{shan2022p} and GLA-GCN \cite{yu2023gla} in terms of accuracy. When compared to larger models such as MotionAGFormer \cite{mehraban2024motionagformer} and MotionBERT \cite{zhu2023motionbert}, our method exhibits slightly lower computational complexity and faster inference speed. The main reduction in computation for our method is due to a simpler approach to constructing the adjacency matrix, as opposed to the K-nearest neighbor method used in the temporal GCN module in \cite{mehraban2024motionagformer}.

These experimental results demonstrate that our model possesses both robustness and efficiency.

\subsection{Qualitative results}

\begin{figure}[htbp]
  \centering
  \includegraphics[scale=0.35]{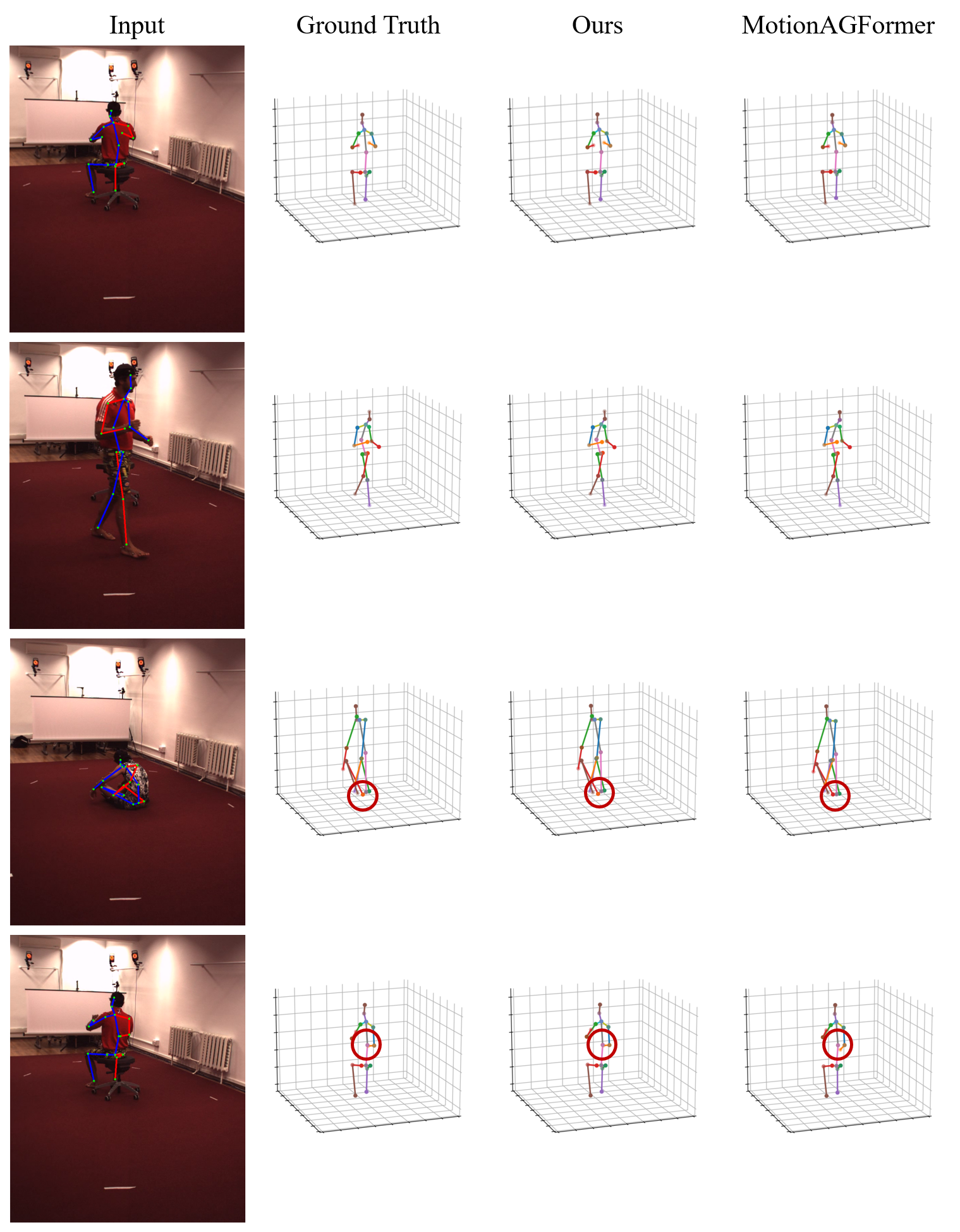}
  \caption{Pose estimation visualization results for Human3.6M.}
  \label{fig:fig4}
\end{figure}

For the Human3.6M dataset, we conducted 3D human pose reconstruction experiments on both the training and test action videos. We present some visualization results in Figure \ref{fig:fig4}, including actions such as Sitting, Walking, Eating, and Sitting Down. We compared our reconstruction results with the ground truth and SOTA method. Our estimated poses are highly accurate for actions involving rich information, such as Walking and Eating. For more challenging actions, such as Sitting and Sitting Down, the visualization results demonstrate that most of the generated poses are highly impressive and remarkably close to the ground truth. In some details, our model even outperforms the baseline model \cite{mehraban2024motionagformer}.

\begin{figure}[H]
  \centering
  \includegraphics[scale=0.37]{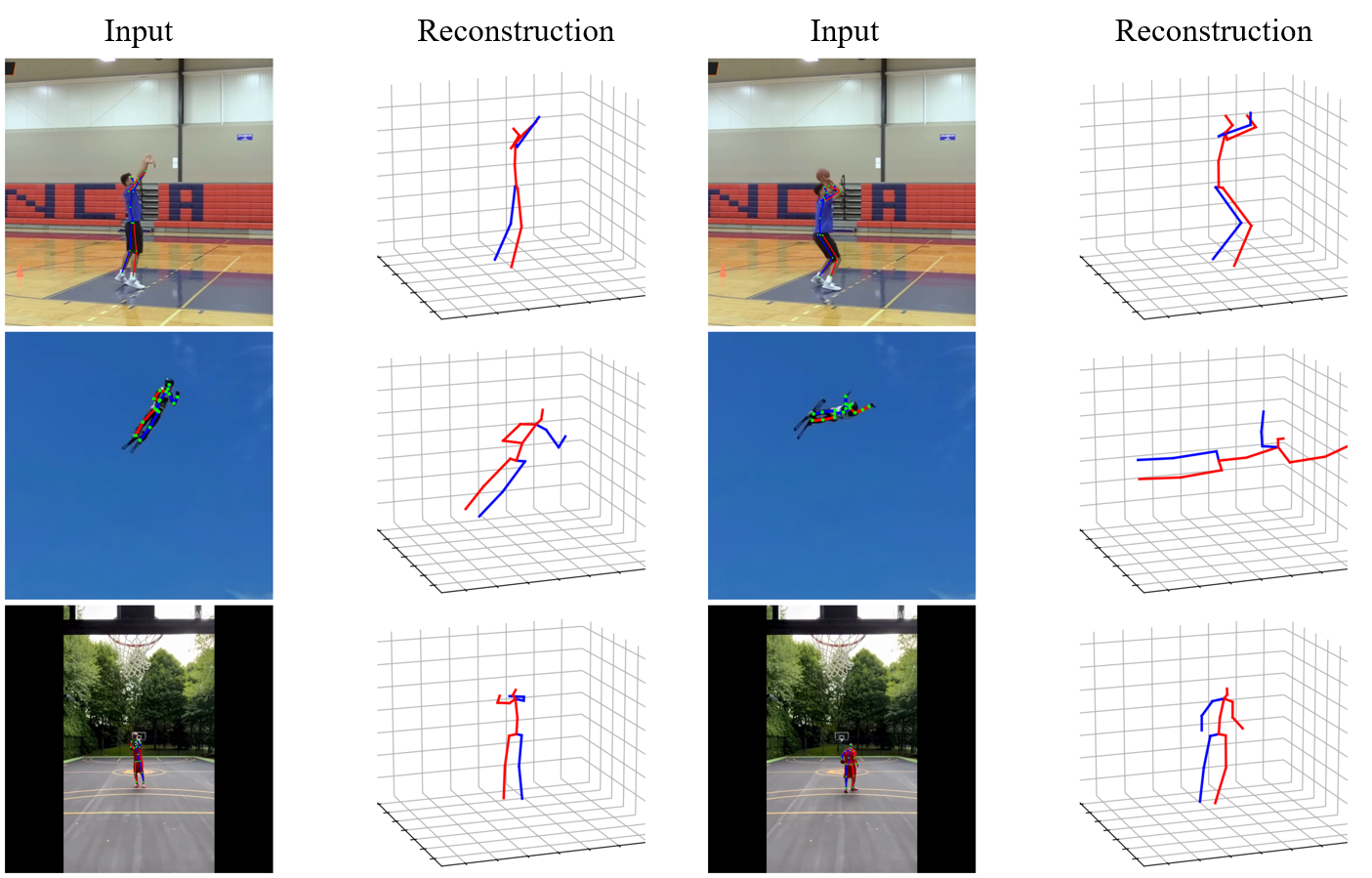}
  \caption{Pose estimation visualization for in-the-wild videos.}
  \label{fig:fig5}
\end{figure}

Our model also performs effectively on in-the-wild videos. Using our fine-tuned model, we conducted 3D pose estimation on these videos. The specific visualization results are presented in Figure \ref{fig:fig5}. We provide four sets of action video examples, including complex actions such as playing basketball, freestyle skiing, and dancing.

To verify the performance and generalization of our pre-trained model in different scenarios, we applied it to the task of human action recognition. Human action recognition leverages machine learning and deep learning technologies to identify and understand human actions, and it is widely used in security monitoring, sports analysis, and human-computer interaction. Among various methods, skeleton-based human action recognition is particularly important, primarily utilizing 2D information of the human skeleton to identify and classify human actions.

Similar to the two-stage HPE, skeleton-based human action recognition also involves obtaining the 2D posture of the human skeleton, which can be extracted using the 2D HPE algorithm. By analyzing the features in the temporal sequence and the spatial relationships of the joints, the model can identify different types of actions. Mainstream skeleton-based action recognition and 3D human pose estimation techniques are similar in focusing on temporal and spatial modeling, with deep learning models such as LSTM, GCN, CNN, TCN, and Transformer commonly used to tackle this problem. \cite{cheng2020skeleton, duan2022revisiting, chen2021channel, zhu2023motionbert, li20213d}.

\begin{figure*}[!t]
  \centering
  \includegraphics[scale=0.58]{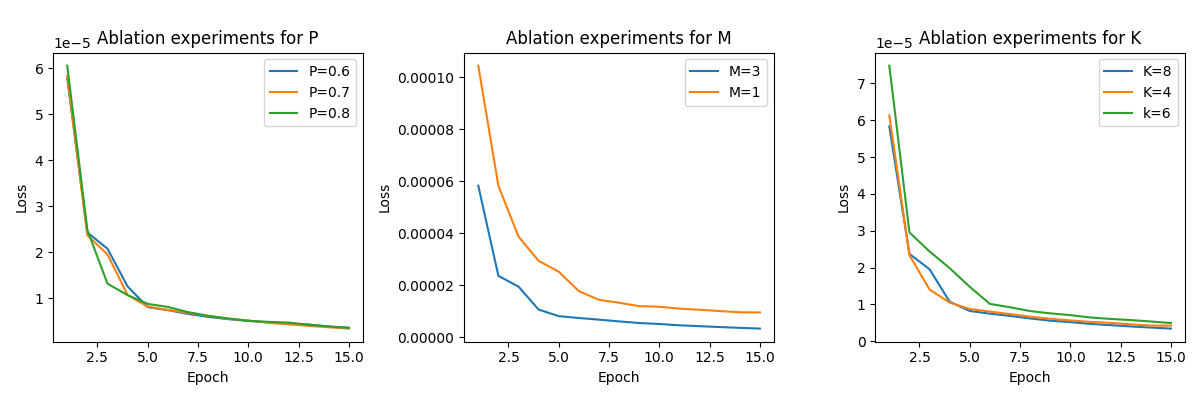}
  \captionsetup{justification=centering}
  \caption{Ablation training results in the pre-training stage.}
  \label{fig:fig6}
\end{figure*}

For the model structure, our human action recognition model is essentially the same as the one introduced in the previous section, with only minor modifications. In the 3D pose estimation task, we use a projection head to map high-dimensional features into the 3D pose space. For human action recognition, we perform global average pooling on the temporal-spatial features output by the network backbone and then send the result to the classification head. The classification head is essentially a MLP or a linear layer with hidden layers. Regarding training targets, the 3D pose estimation task primarily utilizes the L2 loss, whereas the human action recognition task employs the cross-entropy classification loss.

For experimental data, we use the NTU-RGB+D dataset \cite{shahroudy2016ntu}, which contains 57,000 videos across 60 action categories. Following previous methods \cite{cheng2020skeleton, zhu2023motionbert}, we use two data segmentation forms: cross-subject (X-Sub) and cross-view (X-View). Similar to the 3D pose estimation task, we use HRNet to extract the 2D pose skeletons for network training and testing. The evaluation metrics used are Top-1 accuracy.

For the experimental configuration, we employ the same setup as our base and large model. To verify the generalization of our model and the effectiveness of contextualized representations learning, we conducted comparative experiments with and without the pre-trained model. As shown in Table \ref{tab:table8}, our model successfully transfers to human action recognition and performs better than the baseline model motionBERT \cite{zhu2023motionbert}. The Top-1 accuracy results on the two dataset splits are 88.1\% and 93.4\% for our base model, and are 89.5\% and 93.5\% for our large model. Using the contextualized representations pre-trained model, our base model improves the Top-1 accuracy by 1.1\% and 1.5\% and the large model improves by 1.5\% for both settings,  validating the effectiveness of our contextual feature learning method.

\begin{table}[htbp]
\caption{Experimental Results of Human Action Recognition.}
\label{tab:table8}
\centering
\begin{tabular}{@{}l | cc@{}}
\hline
Method & X-Sub & X-View \\
\hline
MotionBERT\cite{zhu2023motionbert} & 87.6 & 93.4 \\
Ours-B(WO Stage1)                  & 87.0 & 91.9 \\
Ours-B(W Stage1)                   & 88.1 & 93.4 \\
Ours-L(WO Stage1)                  & 88.0 & 93.0 \\
Ours-L(W Stage1)                   & 89.5 & 93.5 \\

\hline
\end{tabular}
\vspace{0.1cm}
\end{table}

Overall, our model performs well on public datasets and online videos, accurately reconstructing a variety of poses, including challenging complex actions and high-speed motions. These experiments demonstrate that our method exhibits excellent robustness and generalization ability.

\subsection{Ablation Studies}

For the pre-training stage, we conducted ablation experiments on three key parameters: the mask probability $P$, the number of multi-mask strategies $M$, and the number of pre-training learning target layers $K$. To ensure a fair evaluation, we kept the training epoch fixed at 15 and maintained consistent training conditions, altering only the values of $P$, $M$, and $K$.

As illustrated in Figure \ref{fig:fig6}, we observe significant differences in loss and convergence speed based on the number of multi-mask strategies $M$. When $M=3$, the initial loss is approximately twice that of $M=1$, and the final loss is also significantly smaller compared to when the multi-mask strategy is not used. Regarding the number of pre-training learning target layers, as shown in the right figure of Figure \ref{fig:fig6}, the case of $K=8$ outperforms those of $K=4$ and $K=6$. Specifically, with $K=8$, the initial loss is approximately $5.831 \times 10^{-5}$, and the loss at the end of training is about $3.425 \times 10^{-6}$. In comparison, with $K=4$ and $K=6$, the initial losses are approximately $6.131 \times 10^{-5}$ and $7.480 \times 10^{-5}$, respectively, and the final losses are $4.217 \times 10^{-6}$ and $4.927 \times 10^{-6}$, respectively. Thus, the training results with $K=8$ are superior.

For the mask probability $P$, the differences are not as evident during the pre-training stage alone. Therefore, we controlled the conditions and conducted fine-tuning experiments. We set the pre-training parameters to $M=3$ and $K=8$, then used $P=0.6$, $P=0.7$, and $P=0.8$ to construct the pre-training models and trained them for 60 epochs. Similar methods were used to control variables for other parameters during ablation experiments. The results obtained are presented in Table \ref{tab:table4}.

\begin{table}[htbp]
\centering
\caption{Ablation Experiments for Contextualized Representations Learning.}
\label{tab:table4}
\begin{tabular}{@{}ccc | cc@{}}
\hline
P   & M & K & MPJPE & P-MPJPE \\
\hline
0.8 & 3 & 8 & 38.5  & 32.4    \\
0.7 & 3 & 8 & 38.4  & 32.1    \\
0.6 & 3 & 8 & 39.7  & 33.4    \\
0.7 & 1 & 8 & 38.8  & 32.6    \\
0.7 & 3 & 6 & 38.9  & 32.7    \\
0.7 & 3 & 4 & 39.6  & 33.1    \\
\hline
\end{tabular}
\vspace{0.1cm}
\end{table}

From Table \ref{tab:table4}, we can see that when $P=0.7$, $M=3$, and $K=8$, our fine-tuning results achieve the best performance with an MPJPE of 38.4 mm and a P-MPJPE of 32.1 mm. For the mask probability $P$, the fine-tuning results are very promising when $P=0.7$ and $P=0.8$. Our findings also partially confirm the conclusions of \cite{he2022masked}. Intuitively, pose sequences are similar to images in that they contain large spatial-temporal information redundancy. When feature modeling is performed, using a higher mask rate and creating a challenging learning task can yield good results.

Regarding the number of multi-mask strategies $M$, due to computing power limitations, we could not perform pre-training experiments with larger $M$. However, our results indicate that when $M$ increases from 1 to 3, the MPJPE improves by 0.4 mm and the P-MPJPE improves by 0.5 mm, demonstrating the effectiveness of the multi-mask strategy. For the number of pre-training learning target layers $K$, we find that adjusting and increasing the value of $K$ significantly improves the fine-tuning performance. This suggests that using multi-layer features as pre-training learning targets can not only capture rich contextual features but also stabilize the training process and enhance the model's robustness.

For the configuration of the Transformer-GCN dual-stream model, we conducted ablation experiments on the model parameters, the training configurations, and the modules.

\begin{table}[htbp]
\centering
\caption{Ablation Experiments for Parameter Selections.}
\label{tab:table5}
\begin{tabular}{@{}cc | cc@{}}
\hline
N  & d   & MPJPE & P-MPJPE \\
\hline
6  & 128 & 42.2  & 34.8    \\
8  & 128 & 40.0  & 33.3    \\
12 & 128 & 39.4  & 32.9    \\
16 & 128 & 38.4  & 32.1    \\
20 & 128 & 38.2  & 32.3    \\
16 & 256 & 38.0  & 31.9    \\
\hline
\end{tabular}
\vspace{0.1cm}
\end{table}

For the model parameters, we conducted comparative experiments for network layers $N$ and feature dimension $d$, as shown in Table \ref{tab:table5}. From the comparison of experimental results, we can observe that increasing the scale of the model, such as the number of layers or the hidden layer feature dimension, only slightly improves the model's performance but introduces a computational burden. When $N=16$ and $d=128$, our base model achieves ideal accuracy, with an MPJPE of 38.4 mm and a P-MPJPE of 32.1 mm. When $N=16$ and $d=256$, our large model achieves the best accuracy with an MPJPE of 38.0 mm and a P-MPJPE of 31.9 mm.

For training parameters, we primarily conducted ablation experiments on the selection of the hyperparameter for N-MPJPE $\lambda_1$ and the hyperparameter for the mean per joint velocity error $\lambda_2$. The detailed experimental results are presented in Table \ref{tab:table6}. The results indicate that using only one regularization term, either $\lambda_1$ or $\lambda_2$, leads to suboptimal performance. We conclude that our results achieve the best performance when $\lambda_1=0.5$ and $\lambda_2=20$, demonstrating that the model benefits from balancing the regularization parameters rather than relying on just one.

\begin{table}[htbp]
\centering
\caption{Ablation Experiments for Hyperparameters.}
\label{tab:table6}
\begin{tabular}{@{}cc | cc@{}}
\hline
${\lambda_1}$  & ${\lambda_2}$  & MPJPE & P-MPJPE \\
\hline
0.5 & 20 & 38.4  & 32.1    \\
0   & 0  & 42.1  & 34.1    \\
0.5 & 0  & 39.9  & 33.6    \\
1   & 0  & 42.0  & 35.3    \\
2   & 0  & 42.1  & 35.0    \\
0   & 1  & 41.1  & 34.4    \\
0   & 5  & 39.7  & 33.3    \\
0   & 10 & 39.9  & 33.3    \\
0   & 20 & 39.4  & 32.9    \\
0   & 30 & 39.5  & 32.6    \\
0.5 & 1  & 39.0  & 33.2    \\
0.5 & 10 & 38.9  & 33.0    \\
0   & 20 & 38.5  & 31.9    \\
1   & 20 & 38.7  & 32.1    \\
2   & 20 & 38.7  & 32.2    \\
\hline
\end{tabular}
\vspace{0.1cm}
\end{table}

For the ablation studies on the modules, we conducted experiments on the contextualized representations learning method, spatial and temporal position embedding, the Transformer stream, and the GCN stream. Specifically, \textit{w/wo Stage1} indicates whether contextualized representations learning is used, \textit{TPE} indicates temporal position embedding, \textit{SPE} indicates spatial position embedding, \textit{AF} indicates adaptive fusion method, \textit{SF} indicates summation-based fusion method. As shown in Table \ref{tab:table7}, we observe that using contextualized representations learning improves the model's MPJPE by 0.2 mm and P-MPJPE by 0.4 mm, demonstrating the effectiveness of our pre-training method.

Regarding position embedding, the results show that using only spatial position encoding yields better performance than using only temporal position encoding or using both simultaneously. This indicates that, given the existence of the adjacency matrix in the GCN stream, temporal position embedding can be omitted. Utilizing only spatial position encoding ensures accuracy while reducing computational complexity.

For the Transformer stream and the GCN stream, we find that both streams are essential. The absence of either stream significantly reduces accuracy, which underscores the necessity of considering both global and local interactions in spatial-temporal balanced modeling.

\begin{table}[htbp]
\centering
\caption{Ablation Experiments for Modules.}
\label{tab:table7}
\begin{tabular}{@{}ccccccc | cc@{}}
\hline
w/wo S1 & TPE & SPE & AF & SF & TrS & GS & MPJPE & P-MPJPE \\
\hline
\checkmark           & ×                  & \checkmark   & \checkmark & × & \checkmark           & \checkmark   & 38.4  & 32.1    \\
×                    & ×                  & \checkmark   & \checkmark & × & \checkmark           & \checkmark   & 38.6  & 32.6    \\
\checkmark           & \checkmark         & ×            & \checkmark & × & \checkmark           &\checkmark    & 39.0  & 32.5    \\
\checkmark           & \checkmark         & \checkmark   & \checkmark & × & \checkmark           & \checkmark   & 39.0  & 32.3    \\
\checkmark           & ×                  & ×            & \checkmark & × & \checkmark           & \checkmark   & 40.1  & 33.9    \\ 
\checkmark           & ×                  & \checkmark   & × & \checkmark & \checkmark           & \checkmark   & 40.0  & 33.3    \\
\checkmark           & ×                  & \checkmark   & \checkmark & × & \checkmark           & ×            & 40.3  & 33.3    \\
\checkmark           & ×                  & \checkmark   & \checkmark & × & ×                    & \checkmark   & 40.4  & 33.7    \\
\hline
\end{tabular}
\vspace{0.1cm}
\end{table}

\section{Conclusion}

This paper investigates the Transformer-GCN model with contextualized representations learning for 3D HPE. It integrates 3D HPE technologies, self-attention Transformer, GCN, and contextualized representations learning methods. We utilize a Transformer-GCN dual-stream model that employs masked input features for contextualized representation learning and spatial-temporal modeling, enabling the model to capture spatial-temporal relationships between postures and improve generalization. The dual-stream approach balances global and local interactions by combining the GCN stream, which focuses on local relationships between adjacent key points and frames, with the Transformer stream, which captures global spatial and temporal features. Our model achieves SOTA performance on two benchmarks, Human3.6M and MPI-INF-3DHP, and performs well on both public datasets and online videos.

\textbf{Discussion} Although our work achieves SOTA performance among pre-training methods for 3D HPE, some diffusion-based methods were not included in the comparisons, primarily due to their unaffordable computational complexity. Diffusion-based methods \cite{shan2023diffusion, zhou2023diff3dhpe, cai2024disentangled, gong2023diffpose} typically add Gaussian noise to real 3D poses and then use a denoiser to refine the 3D joint positions sampled from the noisy data. Specifically, these models can produce more accurate 3D poses at each denoising step, while the iterative denoising process allows the noisy pose sequences to gradually converge to the ground truth. However, such methods require substantial computational resources, as the iterative denoising process is time-consuming both during inference. For example, in \cite{shan2023diffusion}, when the number of hypotheses is set to 20 and the number of iterations is set to 10, the MACs are approximately 200 times higher than when the number of hypotheses is set to 1, reaching 228.2G. In contrast, our large model achieves superior accuracy with a computational cost of only 91.8G.

\textbf{Future Work} The common challenge in 2D-to-3D HPE techniques is its reliance on the accuracy of preliminary 2D pose data, since errors in this foundational 2D pose data can negatively impact subsequent 3D estimations. Meanwhile, most video-based pose estimation models impose constraints on input dimensions, such as 17 keypoints used in most works. Exploring approaches to eliminate these restrictions and enable unconstrained data input for model training represents a promising future direction. Such advancements would allow for the simultaneous training of data with varying frame counts and keypoint distribution patterns, facilitating diverse 3D pose outputs. This approach holds greater practical application potential, enabling models to achieve a deeper understanding of pose estimation.  Our method leverages a Transformer-GCN dual-stream model to learn high-dimensional representations through a self-distillation framework, incorporating masked 2D pose features to enhance learning. By employing the Contextualized Representations Learning approach for feature pre-training, this method enables adaptability to various 2D pose input formats and presents potential for extension to image-based data integration, further broadening its application scope.

\bibliographystyle{unsrt}


\vspace{11pt}


\vfill
\end{document}